# Enhanced Temporal Processing in Spiking Neural Networks for Static Object Detection Using 3D Convolutions

Huaxu He[1]

School of Computer and Information Engineering, Henan University[1]

E-mail: 104753230975@henu.edu.cn

## Abstract

Spiking Neural Networks (SNNs) are a class of network models capable of processing spatiotemporal information, with event-driven characteristics and energy efficiency advantages. Recently, directly trained SNNs have shown potential to match or surpass the performance of traditional Artificial Neural Networks (ANNs) in classification tasks. However, in object detection tasks, directly trained SNNs still exhibit a significant performance gap compared to ANNs when tested on frame-based static object datasets (such as COCO2017). Therefore, bridging this performance gap and enabling directly trained SNNs to achieve performance comparable to ANNs on these static datasets has become one of the key challenges in the development of SNNs.To address this challenge, this paper focuses on enhancing the SNN's unique ability to process spatiotemporal information. Spiking neurons, as the core components of SNNs, facilitate the exchange of information between different temporal channels during the process of converting input floating-point data into binary spike signals. However, existing neuron models still have certain limitations in the communication of temporal information. Some studies have even suggested that disabling the backpropagation in the time dimension during SNN training can still yield good training results. To improve the SNN handling of temporal information, this paper proposes replacing traditional 2D convolutions with 3D convolutions, thus directly incorporating temporal information into the convolutional process. Additionally, temporal information recurrence mechanism is introduced within the neurons to further enhance the neurons' efficiency in utilizing temporal information.Experimental results show that the proposed method enables directly trained SNNs to achieve performance levels comparable to ANNs on the COCO2017 and VOC datasets. The code for this job can be contacted via email.

## 1  Introduction

Spiking Neural Networks (SNNs), as the third generation of artificial neural networks[1]，have demonstrated significant advantages over traditional Artificial Neural Networks (ANNs) in terms of biological interpretability and low energy consumption. These advantages make SNNs highly promising for applications in intelligent computing, edge computing, and other fields[2].The event-driven nature of SNNs enables extremely high energy efficiency in deployment on neuromorphic chips[3].Currently, there are two mainstream training approaches for SNNs: the ANN-to-SNN method [4-8]and direct training of SNNs [9-14].While the ANN-to-SNN approach has shown good results in certain tasks, it typically requires longer time steps to simulate the behavior of ANNs, which not only increases computational complexity but may also be limited by the performance of the original ANN. On the other hand, direct training of SNNs offers advantages such as lower time steps and reduced latency, making it more suitable for real-time processing tasks. This study focuses on optimizing the direct training method for SNNs.

In recent years, significant progress has been made in direct training SNNs in various aspects, including the adoption of advanced network architectures from ANNs (such as Spiking ResNet[15, 16]、Spikformer[17]) and improvements to neuron models (such as CLIF[18]、PLIF[19]、GLIF[20]、DSGM+DTAM[21]).Additionally, other improvements to SNNs （such as SLT[22]、TET[23]、tdBN[24]),have also driven the development of SNNs technology. In classification tasks, SNNs have demonstrated performance comparable to, and in some cases surpassing, that of ANNs on multiple datasets[25]. However, in the classic computer vision task of object detection (i.e., identifying objects and accurately localizing their bounding boxes), SNNs still exhibit a performance gap compared to ANNs. Specifically, although SNNs outperform ANNs on event-based neuromorphic datasets, the performance gap remains significant on traditional static datasets (such as COCO2017 and VOC)[26]. The root of this issue lies in the fact that, although current SNNs possess certain capabilities in processing time-related information, they still fall short in handling and integrating temporal sequence information.In Section 2.4, we will explore the underlying causes of this issue and validate its existence through experiments in Section 4.4.

Rather than adopting ANN-inspired architectures for SNNs, direct optimization of SNNs aims primarily at enhancing or leveraging their unique ability to process temporal information. In SNNs, spiking neurons are the key components for handling spatiotemporal information. However, despite numerous improvements to neuron models in recent years (such as CLIF, PLIF, etc.) [18-20], the temporal information transfer in SNNs remains relatively weak. This lack of temporal communication has led to the dominance of spatial information during gradient propagation in SNNs, which has prompted some researchers to propose training strategies that do not calculate temporal gradients or only calculate partial neural temporal gradients[27-29].To better exploit temporal information, researchers have introduced temporal channel attention mechanisms, which have shown significant results in image generation tasks[30, 31]. However, implementing attention mechanisms typically requires extensive multiplication operations to dynamically adjust attention weights, which can reduce efficiency in practical applications. To address this, the Gated Attention Coding (GAC) method applies the attention mechanism solely to the encoding process in the first layer of the network, thereby enhancing performance in classification tasks[32]. Nevertheless, this method still yields limited performance improvements in object detection tasks.

To effectively transmit temporal information, this paper proposes a straightforward enhancement—replacing the 2D convolution in the network with 3D convolution. This change allows temporal information to be processed alongside the spatial dimensions (i.e., width and height of the input image), enabling direct integration of spatiotemporal information. Through this approach, communication between different temporal channels becomes more direct and efficient. Additionally, this study combines neuron models from [33] and [34], further enhancing the network's ability to process temporal channel information, and introduces a temporal information recurrence mechanism within the neurons to improve their efficiency in utilizing temporal information.The experimental results show that the proposed method achieves performance comparable to traditional ANN models on the COCO2017[35] and VOC[36] datasets, and verifies the effectiveness of the method in enhancing the processing of temporal information.

The main contributions of this paper are summarized as follows：

- To enable the network to adopt differentiated processing strategies for input data with varying temporal steps, we propose replacing the 2D convolution kernel in traditional SNNs with a 3D convolution kernel. By directly integrating temporal information into the convolutional process, the network's ability to process temporal information is significantly enhanced.

- A new membrane potential initialization method is introduced, incorporating a temporal information recurrence mechanism. This breaks the traditional unidirectional transmission of temporal information, thereby improving the network's efficiency in utilizing temporal information.

- The effectiveness of the proposed methods has been validated through experiments. The experimental results show that the proposed approach enables the SNN to achieve performance that matches or even surpasses that of an ANN with the same architecture. Ablation studies and generalization experiments confirm the robustness and adaptability of the method, while temporal enhancement validation experiments further demonstrate its effectiveness in improving the SNN's ability to process temporal information.

## 2 Related Works

### 2.1 Deep Spiking Neural Networks

Currently, the mainstream deep SNN training approaches can be categorized into two primary strategies: converting ANNs to SNNs and directly training SNNs. The former approximates the behavior of an ANN by simulating the average spike rate in the SNN, with its performance heavily dependent on the original ANN. However, this method often encounters performance degradation and high latency during the conversion process. In contrast, direct training faced early challenges due to the non-differentiability of the spike function, which made it difficult to apply traditional backpropagation algorithms. With the introduction of surrogate functions and the development of the Spatio-Temporal Backpropagation (STBP) method, direct training has gradually become one of the mainstream strategies for SNN training. This paper focuses on optimizing and experimenting with direct training methods for SNNs.

Deep residual structures are critical for enhancing network performance. In the context of directly trained SNNs, there are two main approaches to residual design: SEW-ResNet accumulates the 0/1 spike signals generated by the neurons, while MS-ResNet accumulates the input currents (floating-point data) before the neurons. Both approaches enable the effective training of neural networks with more than 100 layers[15, 16]. In this study, the SEW-ResNet linking approach is primarily adopted for network implementation.

### 2.2 Low-Energy Object Detection

Object detection tasks can be categorized into static object detection and event-based object detection, which are typically performed using frame-based static datasets and neuromorphic datasets from event cameras, respectively. The former refers to 2D images captured by standard cameras, while the latter captures intensity variations asynchronously and sparsely using biomimetic sensors, producing sparse event streams with high temporal resolution, low latency, and low power consumption, making it suitable for complex scene processing[37]. This study focuses on static object detection tasks.

The YOLO (You Only Look Once) series of models are among the most widely used network architectures in object detection tasks [38]. Their ANN-based implementations have demonstrated outstanding performance in both academic and industrial domains, leading to their widespread adoption. However, high performance is often accompanied by high energy consumption. In contrast, spiking neural networks (SNNs), with their low-energy characteristics, offer a potential solution for improving the energy efficiency of the YOLO model.

Object detection based on SNNs remains a significant challenge, with limited research conducted in this area, particularly on directly trained SNNs. Early spiking YOLO models primarily relied on the ANN-to-SNN conversion approach [39-41]. However, this approach has several limitations, including performance ceilings constrained by the original ANN model, high latency due to long time steps, and poor adaptability to event cameras. To address these challenges, the direct training-based EMS-YOLO model was introduced. While direct training-based EMS-YOLO has shown remarkable success on event-based datasets, its performance on frame-based static datasets still lags behind that of ANN models [26, 42]. In this work, we present the first successful application of a spiking YOLO model trained on a static dataset, achieving performance comparable to that of the same-structured ANN.

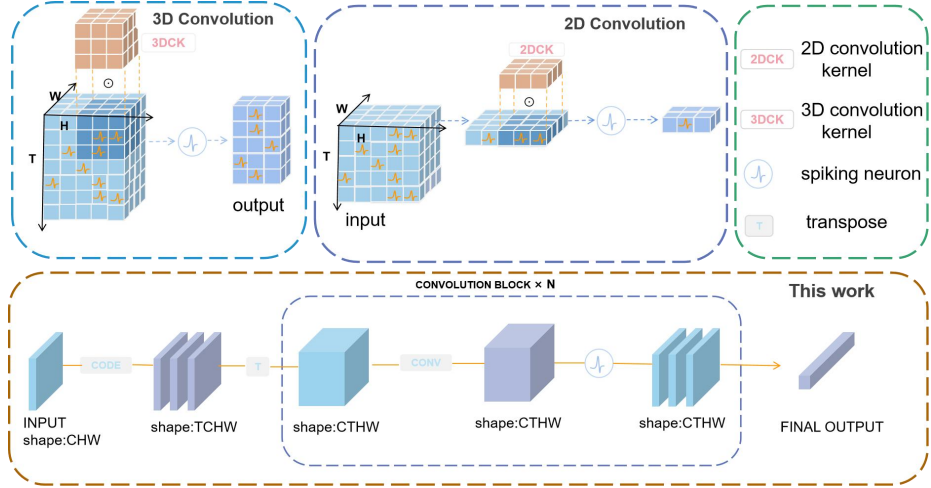

Figure 1, Overall Structure Diagram

### 2.3 Limitations of Temporal Processing in SNNs

A potential issue in Spiking Neural Networks (SNNs) is that, although spiking neurons provide limited spatiotemporal information processing capabilities, the additional temporal dimension introduced in the network significantly increases the complexity of training. A study in [43] revealed that inputs from different time steps lead to variations in output distributions, causing interference from different optimization directions during the training process, particularly when using neuromorphic datasets. To address this issue, the study employed the Kullback-Leibler (KL) divergence method to make the output distributions of different time steps more similar, which effectively improved the network's performance and reduced the number of time steps required for training.

In another study, [33] pointed out that inputs from different time steps have distinct meanings and levels of importance, especially when using temporal coding. Consequently, the network needs to adopt differentiated processing strategies based on the data characteristics of each time step and selectively discard certain time step inputs. This study introduced flexible, learnable neuron membrane potentials, enabling the network to have distinct learnable parameters at different time steps, thereby enhancing performance and making temporal coding superior to direct coding. However, both of these approaches still rely on traditional architectures, using the same convolutional kernels to process information from different time steps, which does not fundamentally solve the issue.

### 2.4 Encoding Schemes

Spike encoding aims to transform input floating-point data into spiking signals represented by 0/1 spiking with a temporal dimension. Directly trained SNNs typically employ direct encoding[34, 44, 45], where a convolutional layer processes the input image, and the feature map is repeated across multiple time steps before being fed into the neurons, generating the spiking signals required by the SNN. However, this method uses the same feature map at each time step, failing to fully exploit the spatiotemporal properties of the SNN. In contrast, the time to first spike coding generates spike sequences with spatiotemporal characteristics and reduces energy consumption, although it may lose significant amounts of original data at low time steps[46-48].

To address the limitations of these two encoding methods, [33] proposed a hybrid scheme that combines direct encoding and time encoding, enhancing the performance of time encoding at low time steps. In the ablation experiments of this study, the performance difference between direct encoding and hybrid time encoding was evaluated, and further improvements were achieved by integrating the GAC mechanism [32], thereby enhancing the encoding capability and optimizing the overall network performance.

## 3 Methods

### 3.1 3D Convolution:

Early researchers tended to approximate SNNs as recurrent neural networks (RNNs) driven by binary spikes, primarily due to the fact that the states of spiking neurons in SNNs are tightly coupled across different time steps [14, 49]. However, compared to RNNs, SNNs are more aligned with the original neural network structure, which requires handling additional temporal dimensions in the data.

In SNNs, after encoding an image, the original 3D floating-point data in the CHW (channel-height-width) format is transformed into 4D binary spike data in the TCHW (timestep-channel-height-width) format. This transition from ANN involves two key changes: first, the data dimensions expand from three to four, and second, the data type shifts from floating-point to binary spikes. In the framework of CNNs, a simple and intuitive approach is to replace the 2D convolution used for processing 3D data with 3D convolutions that can handle 4D data. As shown in Figure 1, network residual connections are omitted for clarity.

When processing output images with a 2D convolution kernel, the network first encodes the input CHW format data into four-dimensional TCHW or CTHW data containing the time dimension. Subsequently, the network performs multiple convolution operations and spiking neuron processing based on depth, with the data format consistently maintained as TCHW or CTHW. When using a 3D convolution kernel and the encoded data is TCHW, the image must first be transposed, placing the channel dimension (C) at the front to meet the processing requirements of the 3D convolution. Afterward, the network performs multiple convolutions and spiking neuron processing, with the data format consistently maintained as CTHW.

In the convolution process, when using a 2D convolution kernel, the network performs T independent 2D convolution operations on the TCHW format data, with the CHW data at each time step processed by the same 2D convolution kernel. This approach not only limits the communication of information between temporal channels but also causes the same convolution kernel to be applied across different temporal channels, thereby weakening the spatiotemporal information processing capability of the SNN. In contrast, when using a 3D convolution kernel, the CTHW formatted data is processed in a single 3D convolution operation, enabling direct and effective communication of information between channels.

It is important to note that even with 3D convolutions enhancing temporal information exchange, the spiking neurons still play a crucial role. Neurons not only provide essential nonlinear functions in the network but also coordinate the information exchange between temporal channels over long time steps.

### 3.2 Spiking Neuron Model

In SNNs, the function of neurons is similar to activation functions in ANNs, as they convert input floating-point signals into discrete 0/1 spike signals. The design of neurons can vary in complexity, but overly complex models are often constrained by current hardware limitations, especially when the network depth increases, leading to significant computational overhead and complexity. Therefore, in practical applications, the use of complex neuron models is relatively rare. The commonly used neuron models in SNNs are described by the following equations:

$$H[t] = f(V[t-1], X[t]) \qquad (1)$$

$$S[t] = \Theta(H[t] - V_{th}) \qquad (2)$$

$$V[t] = H[t](1 - S[t]) + V_{reset}S[t] \qquad (3)$$

Where X[t] represents the input current at time step t. When the membrane potential exceeds a threshold value, the neuron fires a spike. Θ(x) is the Heaviside step function, which equals 1 for x⩾0 and 0 otherwise. V[t] denotes the membrane potential after a spiking event; if no spike is generated, it remains equal to H[t], otherwise, it is set to the reset potential. The functional descriptions of the IF[50] and LIF[51, 52] models can be formulated in Equations (4) and (5), respectively, as follows:

$$H[t] = V[t-1] + X[t] \qquad (4)$$

$$H[t] = V[t-1] + \frac{1}{\tau}(X[t] - (V[t-1] - V_{reset})) \qquad (5)$$

Among them, τ represents the membrane time constant. Equations (2) and (3) describe the generation and resetting process of spikes, which are consistent across all types of spiking neuron models.

The type of neuron plays a crucial role in network performance. Classic LIF neurons are computationally simple but often do not achieve optimal performance. In contrast, more complex neuron models can more accurately simulate biological intelligence systems, though they are often constrained by hardware limitations. To address this, we balance network performance and training resource consumption by combining neuron models from [33] and [34], resulting in significant performance improvements. In the modified neuron model, the membrane potential and input current are combined as shown in equation (6):

$$H[t] = l^t V[t-1] + i^t X[t] \qquad (6)$$

where $l^t$ and $i^t$ are learnable parameters, representing the membrane potential decay constant and input current constant, respectively. The advantage of this modification is that the neuron's dynamics during spike generation become more complex. When $l^t$>1, the membrane potential shifts from decay to increase; when $l^t$<0, the membrane potential inhibits the next time step's spike. When $l^t$ and $i^t$ are both set to 1, equation (6) degenerates into the IF neuron model; in particular, when $l^t + i^t$ =1 and $l^t$ >0, $i^t$ >0, equation (6) becomes equivalent to the LIF neuron model. For the spike generation and reset mechanism of neurons, equations (2) and (3) can be modified as follows:

$$S[t] = \Theta\left(\frac{H[t]}{V_{re}} - V_{th}\right) \qquad (7)$$

$$V[t] = H[t] - V_{re}S[t] \qquad (8)$$

The parameter $V_{th}$ is generally set to 1, while $V_{re}$ denotes the learnable firing threshold. The advantage of equations (7) and (8) lies in their ability to incorporate a learnable firing threshold within the network. While the modified neuron model incurs a slight increase in computational complexity compared to the traditional LIF model, it results in a notable enhancement in network performance.

### 3.3 Temporal Information Recurrence Mechanism

As shown in Equation (1), in traditional spiking neuron models, information transmission along the temporal dimension is unidirectional, meaning that information flows only from earlier to later time steps. While this characteristic may have its unique biological significance in biological neural systems, it is not a necessary constraint in computer-simulated SNNs. By replacing the 2D convolution kernel with a 3D convolution kernel, effective communication of information across adjacent time steps is achieved, breaking the traditional rule of unidirectional temporal information flow. Building upon this, this paper further optimizes the initialization of membrane potential in spiking neurons, changing the common practice of initializing to zero to using the input current from the last time step as the initialization value. Specifically, information is first propagated backward in time, where the input current from the last time step is passed to the membrane potential of the initial time step.

Subsequently, temporal information flows forward as usual, thereby forming an internal information recurrence mechanism within the neuron. As shown in Figure 2, where $X_n^t$ represents the input current at time step t for the n-th layer, and $V_n^t$ represents the membrane potential at time step t for the n-th layer.

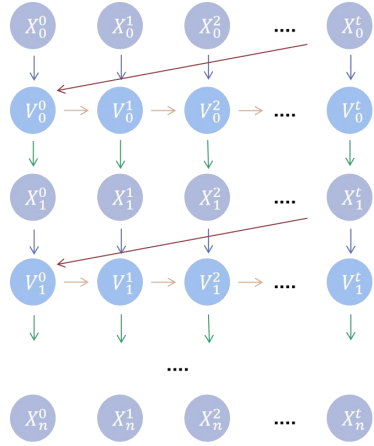

Figure 2: Illustration of the Temporal Information Recurrence Mechanism

By breaking the constraint of unidirectional temporal information flow, neurons can more fully utilize temporal information, allowing information from different time steps to be processed more effectively. This improvement not only enhances the network's capability to handle temporal information but also contributes to better overall performance and generalization of the network.

## 4 Experiments

This section is divided into five parts: the first part introduces the experimental details, the second part discusses the effectiveness experiment, the third part presents the ablation study, the fourth part verifies the generalization ability, and the fifth part examines temporal enhancement validation.

### 4.1 Experimental Details

The key details of the experiment are as follows: The experiment is conducted using the PyTorch framework, with the YOLOv5[52] network architecture as the base, which has been structurally modified. Specifically, the activation functions in the network are replaced with spiking neurons, and introduced residual connections in the form of SEW ResNet[16] commonly used in SNNs, This connection method propagates information by accumulating the 0/1 spike signals processed by neurons. For the LIF neurons, the value of τ is set to 2.

To accommodate the performance limitations of the experimental hardware, the input image size is set to 224x224 pixels. The performance of the experimental results is evaluated using the mean Average Precision (mAP), specifically mAP@0.5, computed at an Intersection over Union (IoU) threshold of 0.5. The AdamW optimizer is used. The batch size varies depending on the network's time steps: when the time step is 15, the batch size is set to 20; when the time step is smaller than 15, the batch size is set to 32. It is worth noting that, compared to traditional ANNs, training SNNs typically requires more iterations to converge. Therefore, the number of epochs in the experiment is set to 300. The datasets used for this experiment include the widely used COCO2017 dataset and the PASCAL VOC dataset, both of which are standard datasets in international computer vision challenges.

### 4.2 Effectiveness Experiment

In the effectiveness experiment, the network performance is evaluated by comparing the performance gap between SNN using different methods and ANN networks with the same architecture. A smaller performance gap indicates that the performance of the SNN network is closer to that of the ANN network, which reflects better performance, rather than simply comparing the numerical values of mAP@0.5. In this study, YOLOv5n is chosen as the baseline network architecture.

The experimental results show that the proposed method achieves comparable mAP@0.5 scores on both the VOC and COCO2017 datasets when compared to ANN networks. Specifically, when comparing a traditional SNN with LIF neurons and a 2D convolution kernel to an ANN with the same architecture, the SNN has a lower mAP@0.5 by 0.078 on the VOC dataset. This result is consistent with the finding in the EMS-YOLO study, where the mAP@0.5 of SNN is 0.064 lower than that of ANN on the COCO2017 dataset, further validating the performance gap between traditional SNNs and ANNs in object detection tasks.

In contrast, the method proposed in this study reduces the mAP@0.5 gap to only 0.008 on the VOC dataset and further narrows it to 0.001 on the COCO2017 dataset, effectively bridging the performance gap between SNNs and ANNs. The specific data are shown in Table 1. It is worth noting that the model used in [26] employs ResNet34 with an input size of 640x640.

Table 1: Effectiveness Experiment Results. $^{+}$denotes the traditional SNN using LIF neurons and 2D convolutions. $^{*}$indicates that the network was trained for 400 epochs instead of the conventional 300 epochs.

| Dataset | method | Architecture | Coding | Time steps | mAP@0.5 |
|---|---|---|---|---|---|
| VOC | SNN$^{+}$/ANN | YOLOv5n | Direct/- | 15/- | 0.530/0.609 |
|  | This work/ANN | YOLOv5n | Hybrid/- | 15/- | 0.601/0.609 |
|  | This work$^{*}$/ANN | YOLOv5n | Hybrid/- | 15/- | **0.618/0.609** |
| COCO2017 | EMS-YOLO/ANN | ResNet34[26] | direct/- | 4/- | 0.501/0.565 |
|  | This work/ANN | YOLOv5n | Hybrid/- | 15/- | **0.265/0.266** |

In the actual training process, ANN networks typically show no significant improvement in the mAP@0.5 score on the validation set after approximately 120 epochs, whereas the mAP@0.5 of the SNN network continues to increase throughout 300 epochs, as shown in Figure 3. Figures 3(a) and 3(b) illustrate the changes in mAP@0.5 for both SNN and ANN networks as the number of training epochs progresses. The need for more training epochs to converge is a common characteristic of SNN networks, a behavior observed across various convolution kernels, spiking neuron models, and time step settings. Thus, a higher number of training epochs contributes to better performance for SNN networks. When the number of training epochs increased from 300 to 400, the proposed method achieved an mAP@0.5 of 0.618 on the VOC dataset, surpassing the 0.609 achieved by the ANN network.

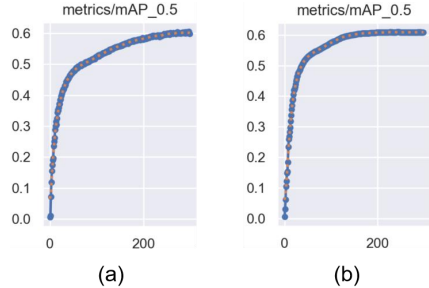

Figure 3: Change in mAP@0.5 on the test set over training epochs.

## 4.3  Ablation Study

In the ablation study, the VOC dataset was used. The experiment began with a subtraction of network components at low time steps to explore the impact of several key factors in the network structure on performance, including whether 3D convolutions were used, whether neurons employed the temporal information recurrence mechanism, and whether GAC was introduced. Compared to the validity experiments, the time step was set to 4, and direct encoding was used. First, when the convolution kernel was changed from 3D to 2D, the mAP@0.5 dropped from 0.567 to 0.534, indicating that 3D convolution has a positive effect on improving network performance. Next, when the temporal information recurrence mechanism was removed, the mAP@0.5 value decreased from 0.567 to 0.552, demonstrating the importance of the temporal information recurrence mechanism for network performance. Additionally, the results showed that GAC positively contributed to the network's performance. The results of this low time-step ablation experiment are shown in Table 2.

Table 2: Low time step ablation study results.

| 3D | recurrence | Gac | mAP@0.5 |
|---|---|---|---|
|  |  |  | 0.567 |
| × |  |  | 0.534 |
|  | × |  | 0.552 |
|  |  | × | 0.550 |

Following this, an additive experiment at higher time steps was performed to examine the impact of key components on network performance. Starting with the basic SNN model, the temporal information recurrence mechanism , the neuron model used in this study, and 3D convolution kernels were progressively added, as shown in Table 3.

Table 3: High time step ablation study results.

| recurrence | Neural | 3D | mAP@0.5 |
|---|---|---|---|
|  |  |  | 0.530 |
| √ |  |  | 0.542 |
| √ | √ |  | 0.565 |
| √ | √ | √ | 0.591 |

In the encoding part of the experiment, direct encoding was compared with hybrid encoding. In the hybrid encoding implementation, the time-encoded information was concatenated with the direct-encoded information along the time dimension. Hybrid encoding exhibited certain advantages over direct encoding, and the experimental results are shown in Table 4. Regarding the impact of time step size on network performance, the results indicate that increasing the time step size can effectively improve network performance, as presented in Table 5. To assess the impact of the neuron model independently, the performance of the LIF neuron model and the neuron model used in this study was compared. In the neuron model testing, both neuron types did not employ the temporal information recurrence mechanism, and both used 2D convolution kernels. The experimental data are shown in Table 6.

Table 4: Comparison of two different encoding methods.

| Method | Coding | Time steps | mAP@0.5 |
|---|---|---|---|
| This work | HYBRID | 15 | 0.601 |
|  | DIRECT | 15 | 0.591 |

Table 5: Time step length test results.

| Method | Coding | Time steps | mAP@0.5 |
|---|---|---|---|
| This work | DIRECT | 4 | 0.567 |
|  |  | 6 | 0.580 |
|  |  | 8 | 0.585 |
|  |  | 15 | 0.591 |

Table 6: Neuron model comparison test results.

| Method | Coding | Time steps | mAP@0.5 |
|---|---|---|---|
| This work | DIRECT | 15 | 0.545 |
| LIF | DIRECT | 15 | 0.530 |

### 4.4 Generalizability Validation

In the generalizability validation, the performance of network models with different parameter sizes and input image sizes was evaluated.

For the model parameter size test, two network models, YOLOv5s and YOLOv5n, were compared. Compared to YOLOv5n, YOLOv5s has a larger network scale and more parameters. Under the condition of a fixed six time steps, the performance of SNN and ANN was tested on these two models. The experimental results showed that for

YOLOv5n, at six time steps, the mAP@0.5 of SNN was 0.580, while the mAP@0.5 of ANN was 0.608, with a difference of 0.028. In contrast, for YOLOv5s, the mAP of SNN was 0.676, and the mAP@0.5 of ANN was 0.687, with a difference of only 0.011. The detailed results are shown in Table 7. These results demonstrate that the proposed method is applicable to network models of different scales.

Table 7 Test results for different-sized network models.

| Method | mAP@0.5 |
|---|---|
| This work/ANN | 0.676/0.687 |
| This work/ANN | 0.580/0.609 |

Regarding the test with different input image sizes, the results indicate that variations in input image size did not significantly affect the network's performance. The specific data can be found in Table 8. In the mAP@0.5 column of the table, the first value represents the result of the proposed method, and the second value represents the result of the ANN model.

Table 8 Test results for different input image sizes.

| Size | mAP@0.5 |
|---|---|
| 224 | 0.567/0.608 |
| 256 | 0.592/0.627 |
| 320 | 0.630/0.668 |

### 4.5 Temporal Enhancement Validation

Enhancing the ability of SNNs to process temporal information is the core of this work. While the effectiveness experiments and ablation studies have demonstrated that the proposed method significantly improves model performance, there has been no direct proof that this improvement stems from enhanced temporal information processing capabilities. Therefore, additional experiments were designed to verify that the performance improvement is not solely due to an increase in the number of parameters but also due to the improvement in the network's ability to process temporal information.

The STBP algorithm primarily utilizes rate and temporal information during the backpropagation process in SNNs. Rate information refers to the average firing rate of neurons in each layer, while temporal information is associated with the order of spike emissions and the membrane potential at each time step[53]. Based on this, we designed relevant experiments. The experimental approach is as follows: first, the temporal information in the network is forcibly converted into rate information, after which the network is trained and tested. The greater the performance loss when the network loses temporal information, the stronger the network's ability to process temporal information; conversely, the smaller the performance loss, the weaker the network's ability to handle temporal information.

To convert the network's temporal information into rate information, we achieved this by modifying the neuron model. Specifically, we shuffled the input currents along the temporal dimension, transforming the spike signals passed to the next layer of the network from ordered to unordered, containing only pure rate information.The experimental results are shown in Table 9, where the first item in the mAP@0.5 column represents the results with temporal information, and the second item represents the results without temporal information. In the difference column, the first item indicates the absolute performance loss, while the second item shows the performance loss as a percentage.

Table 9: Temporal Enhancement Validation.

| Number | Convolution | Recurrence | Neural | Time steps | mAP@0.5 | Difference |
|---|---|---|---|---|---|---|
| 1 | 3D | √ | This work | 15 | 0.551/0.523 | 0.028/0.05% |
| 2 | 3D | √ | This work | 4 | 0.518/0.496 | 0.022/0.042% |
| 3 | 2D | √ | This work | 15 | 0.511/0.501 | 0.01/0.019% |
| 4 | 2D | √ | LIF | 15 | 0.497/0.489 | 0.008/0.016% |
| 5 | 2D | √ | This work | 4 | 0.490/0.485 | 0.005/0.01% |

| | 6 | 2D | × | LIF | 15 | 0.489/0.503 | -0.014/-0.028 |

From the experimental results, the following conclusions can be drawn：：1）Experiments 1 vs. 2, 3 vs. 5: With other conditions constant, only the time step size differs. The results show that networks with larger time steps are more sensitive to temporal information. 2）Experiments 1 vs. 3, 2 vs. 5: With other conditions constant, only the convolution kernel differs. The results indicate that the 3D convolution kernel is more reliant on temporal information compared to the 2D convolution kernel. 3）Experiments 4 vs. 6: With other conditions constant, the temporal information recurrence mechanism enhanced the neuron's sensitivity to temporal information. 4）Experiment 6: when using a traditional SNN, the disappearance of temporal information actually led to an improvement in network performance, indicating that the traditional SNN structure has shortcomings or deficiencies in processing temporal information.

## 5 Conclusion

This paper is the first to demonstrate that directly trained SNNs can achieve performance comparable to that of ANNs on static datasets. To address the significant performance gap between directly trained SNNs and ANNs in the field of object detection, this research focuses on enhancing the unique temporal information processing capabilities of SNNs. To improve the SNN's ability to process temporal information, we replaced 2D convolutional kernels with 3D kernels, enabling the direct incorporation of temporal information processing. Additionally, we introduced a temporal information recurrence mechanism in the neurons to facilitate more efficient utilization of information from the previous layer. While the energy efficiency calculation method is similar to that of EMS-YOLO, we did not repeat the energy consumption calculations in this study. However, there are still certain limitations in the proposed method. While replacing the convolutional kernels improves network performance, it also increases the parameter count and computational complexity, which diminishes the inherent energy efficiency advantage of SNNs. Therefore, modifying the shape of the convolutional kernels could be a potential solution in the future. Nevertheless, this paper effectively enhances the overall performance of the network by introducing additional methods of temporal information communication, without altering the event-driven nature of SNNs, thereby fully demonstrating the potential of SNNs.